\documentclass[10pt, conference]{IEEEtran}

\usepackage{algorithm}
\usepackage{algpseudocode}
\usepackage{graphicx}
\usepackage{subcaption}
\usepackage{float}
\usepackage{amsmath}
\usepackage{tikz}
\usetikzlibrary{shapes, arrows.meta, positioning, backgrounds}
\usepackage{enumitem}
\usepackage{lipsum}
\usepackage{multirow}
\usepackage{float}

\begin{document}

\title{Attire-Based Anomaly Detection in Restricted Areas Using YOLOv8 for Enhanced CCTV Security}

\author{
\IEEEauthorblockN{Abdul Aziz A.B}
\IEEEauthorblockA{
Riyadh, Saudi Arabia\\
bb.abdulaziz@icloud.com}
\and
\IEEEauthorblockN{Aindri Bajpai}
\IEEEauthorblockA{
Chennai, India\\
aindri2003@gmail.com}

}
\maketitle

\begin{abstract}
This research introduces an innovative security enhancement approach, employing advanced image analysis and soft computing. The focus is on an intelligent surveillance system that detects unauthorized individuals in restricted areas by analyzing attire.

Traditional security measures face challenges in monitoring unauthorized access. Leveraging YOLOv8, an advanced object detection algorithm, our system identifies authorized personnel based on their attire in CCTV footage. The methodology involves training the YOLOv8 model on a comprehensive dataset of uniform patterns, ensuring precise recognition in specific regions.

Soft computing techniques enhance adaptability to dynamic environments and varying lighting conditions. This research contributes to image analysis and soft computing, providing a sophisticated security solution. Emphasizing uniform-based anomaly detection, it establishes a foundation for robust security systems in restricted areas. The outcomes highlight the potential of YOLOv8-based surveillance in ensuring safety in sensitive locations.
\end{abstract}


\section{Introduction}
In an epoch defined by progressively intricate security challenges, the imperative for pioneering solutions to fortify safety measures has never been more acute. This research embarks on a quest to formulate an innovative approach, amalgamating image analysis and soft computing techniques, with the principal aim of engendering an intelligent surveillance system. This system is conceived with the intrinsic capability to discern unauthorized individuals within highly sensitive and restricted precincts, relying on the distinctiveness of their attire as a singular identifier.

Conventional security methodologies, while efficacious to a certain extent, frequently contend with inherent limitations when tasked with the surveillance and identification of unauthorized access within fortified environments. To surmount this challenge, our research harnesses the cutting-edge capacities of advanced deep learning technology, prominently featuring YOLOv8 (You Only Look Once version 8) – a preeminent object detection algorithm. YOLOv8 endows our system with the prowess to detect and recognize individuals based on their apparel, an indelible trait of authorized personnel within these secure confines. Scrutinizing closed-circuit television (CCTV) footage with meticulous attention, our system distinguishes individuals through their attire and, with remarkable precision, identifies anomalies that may elude conventional surveillance methodologies. This capability not only heightens security alerts but also mitigates the occurrence of false alarms, a recurrent predicament in traditional security systems.

At the heart of our methodology lies the meticulous training of the YOLOv8 model on an extensive dataset of uniform patterns intricately correlated to specific regions within the secured environs. This data-centric approach ensures unparalleled accuracy in recognizing and classifying individuals, thereby facilitating a swift and highly efficient response to potential security breaches.

Furthermore, our research delves into the domain of soft computing techniques, an indispensable facet of our surveillance system's architecture. The integration of these techniques equips the system with the adaptability to thrive in dynamic environments and diverse lighting conditions. This pivotal component not only fortifies our solution for security enhancement but also contributes to the expansive field of image analysis and soft computing applications, marking a significant stride toward the evolution of intelligent surveillance systems.
By according paramount importance to uniform-based anomaly detection through advanced technology, this research delineates the trajectory for the development of robust and efficient security systems, poised to safeguard restricted areas with unwavering efficacy. The outcomes of this study herald the tremendous potential of YOLOv8-based surveillance systems, underscoring their pivotal role in ensuring the safety and security of sensitive locations.

\section{Related Works}

Within the domain of advanced image analysis, a plethora of scholarly pursuits have made substantial contributions to the sphere of object detection. Among these, the You Only Look Once (YOLO) algorithm stands out as a trailblazing solution, fundamentally reshaping the landscape through its integrated, real-time methodology employing a singular neural network. This survey delves into a comprehensive exploration of object detection algorithms, elucidating their utility across diverse applications.

In the seminal work by Redmon \textit{et al.} \cite{redmon2016}, the YOLO paradigm emerged, introducing a unified, real-time object detection approach utilizing a solitary neural network for predicting bounding boxes and class probabilities. Renowned for its exceptional speed, the algorithm processes images at an impressive rate of 45 frames per second. Subsequent iterations, as exemplified in the work of Redmon and Farhadi \cite{redmon2017}, ushered in YOLO9000, a real-time detection system capable of discerning over 9000 object categories. YOLOv2, an enhanced model, exhibited superior performance, surpassing contemporary methods while maintaining significantly accelerated processing. YOLOv3, a refined version introduced by Redmon and Farhadi \cite{redmon2018}, demonstrated heightened accuracy and speed with minimal design modifications. Bochkovskiy \textit{et al.} \cite{bochkovskiy2020} harnessed YOLOv4 for object detection, amalgamating universal features such as Weighted-Residual-Connections and Cross-Stage-Partial-connections, along with innovative elements like Mosaic data augmentation and Drop Block Regularization, achieving optimal speed and accuracy. Tripathi \textit{et al.} \cite{tripathi2022} presented a meticulous overview of diverse object detection techniques grounded in the YOLO paradigm.

The versatility of YOLO finds expression in varied domains. Yang \textit{et al.} \cite{yang2018} showcased a YOLOv3-based face detection system, characterized by reduced detection time and heightened robustness, ensuring elevated accuracy in intricate environments. Ren \textit{et al.} \cite{ren2017} introduced YOLO-PC, a real-time people counting approach employing boundary selection, applicable across smart cities and achieving detection rates exceeding 40 frames per second with GPU support. The utility of YOLO extended to Automated Drone Detection \cite{singha2021}, exhibiting superior performance in comparison to previous studies in drone detection. Ashraf \textit{et al.} \cite{ashraf2022} proposed a YOLO-based framework incorporating Area of Interest (AOI) to mitigate false negatives and false positives in weapon detection systems, a realm where conventional CNN-based systems face computational challenges.

Comparative analyses with region proposal methods and region-based convolutional neural networks (R-CNNs), as exemplified by Ren and He \cite{ren2015}, demonstrated the supremacy of YOLOv2 over Faster R-CNN in terms of mean average precision (mAP) and frames per second (FPS). Integration of ResNet into the feature extraction of the YOLOv3 framework, as demonstrated by Lu \textit{et al.} \cite{lu2019}, further underscored the algorithm's efficacy.

Propounded by Geethapriya \textit{et al.} \cite{geethapriya2019} as a singular neural network algorithm for object detection, YOLO outshone traditional methods such as CNN and R-CNN in terms of efficiency. Li \textit{et al.} \cite{li2022} introduced YOLOv6 as a Single State Object Detection Framework for Industrial Applications, excelling in accuracy and speed through tailored network sizes, self-distillation, and advanced detection techniques. Lou \textit{et al.} \cite{lou2023} enhanced YOLOv8 with the DC-YOLOv8 algorithm, focusing on small-size object detection in intricate scenes.

Collectively, these scholarly works illuminate the diverse research landscape in object detection, offering nuanced insights into various methodologies and techniques while showcasing numerous practical applications.

\section{Methodology}

\subsection{Data Collection and Preprocessing}
Curating a diverse dataset encompassing various attire patterns associated with authorized personnel in restricted areas, employing manual annotation to label images with corresponding attire categories to ensure the model's accuracy during training. Standardizing and augmenting the dataset to enhance model robustness, including resizing, normalization, and data augmentation.

\subsection{YOLOv8 Model Training}
 Detailing the choice of YOLOv8 architecture for its efficiency in real-time object detection as well as leveraging pre-trained models on large datasets to expedite convergence and optimize performance via transfer learning. Hyperparameter tuning iteratively refines model parameters to achieve optimal detection accuracy for attire-based anomalies.
 
\subsection{Soft Computing Integration}
Employing fuzzy logic to enhance the system's adaptability to dynamic environments, accounting for variations in lighting and attire appearances and implementing adaptive thresholding techniques to optimize anomaly detection sensitivity based on changing environmental conditions.

\subsection{Validation and Testing}
Dividing the dataset into training, validation, and testing sets to evaluate the model's performance objectively by employing k-fold cross-validation to assess the model's consistency and generalization across different data subsets. Utilizing precision, recall, F1 score, and mAP (mean Average Precision) to quantify and evaluate the YOLOv8 model's accuracy in detecting attire-based anomalies.

\subsection{System Integration and Deployment}
Integrating the trained YOLOv8 model into a real-time surveillance system for seamless anomaly detection after addressing hardware requirements for efficient deployment, ensuring compatibility with CCTV infrastructure.

\section{Solution Architecture}

Our Solution Architecture consists of two components: Person Detection and Dress Detection. We
created a pipeline where an image is first passed through the Person Detection layer which detects whether a human is present or not. Then, if a human is present, a bounding box for the human is drawn and the image is cropped to showcase only this bounding box. This ensures that the system does not waste time trying to detect dresses when there is no person in the image.
The image is then passed through our dress detector to detect if a particular dress is present in the image. For this particular system we have considered five categories - Jacket, T-Shirt, Shorts, Skirt, Top.

\tikzstyle{block} = [rectangle, draw, text width=5em, text centered, rounded corners, minimum height=3em]
\tikzstyle{arrow} = [->, >=Stealth]

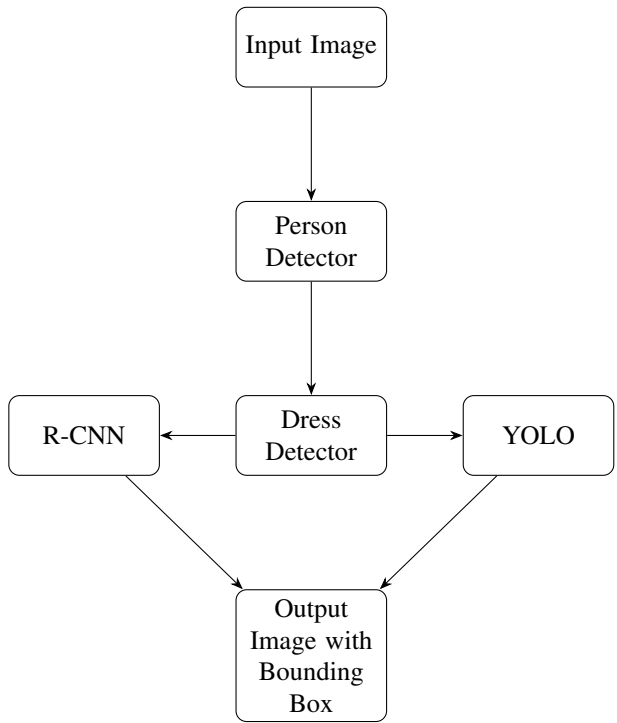
\begin{figure}[!htbp]
\centering
\begin{tikzpicture}[node distance=1.5cm]
    \node (input) [block] {Input Image};
    \node (person) [block, below=of input] {Person Detector};
    \node (dress) [block, below=of person] {Dress Detector};
    \node (yolo) [block, right=of dress, xshift=-0.5cm] {YOLO};
    \node (rcnn) [block, left=of dress, xshift=0.5cm] {R-CNN};
    \node (output) [block, below=of dress] {Output Image with Bounding Box};

    \draw [arrow] (input) -- (person);
    \draw [arrow] (person) -- (dress);
    \draw [arrow] (dress) -- (yolo);
    \draw [arrow] (dress) -- (rcnn);
    \draw [arrow] (yolo) -- (output);
    \draw [arrow] (rcnn) -- (output);
\end{tikzpicture}
\caption{Simplified Workflow Architecture}
\end{figure}

\begin{itemize}
  \item \textbf{Input Image:} The process initiates with the input image, representing the visual data that undergoes subsequent stages of analysis.
  \item \textbf{Preprocessing:} Before diving into the core detection tasks, the input image undergoes preprocessing, involving operations such as normalization, resizing, and other transformations to prepare the image for efficient analysis.
  \item \textbf{Backbone Network:} The preprocessed image is then fed into a \textit{Backbone Network}, a crucial element responsible for feature extraction. This network captures hierarchical features, enabling the subsequent detectors to make informed decisions.
  \item \textbf{Person Detector:} The first specialized detector focuses on identifying individuals within the image. It utilizes the features extracted by the \textit{Backbone Network} to localize and classify persons.
  \item \textbf{Mask Detector:} Simultaneously, the \textit{Mask Detector} operates on the same set of features, aiming to identify the presence of facial masks on the detected individuals. This component is crucial in scenarios where adherence to safety protocols is essential.
  \item \textbf{Dress Detector:} Expanding the scope of object detection, the \textit{Dress Detector} focuses on identifying clothing items within the image. This adds a layer of detail to the analysis, especially relevant in applications such as security or fashion.
  \item \textbf{YOLO (You Only Look Once):} The \textit{Mask Detector} and \textit{Dress Detector} share information with the \textit{YOLO} object detection algorithm. \textit{YOLO}, a state-of-the-art approach, simultaneously predicts bounding boxes and class probabilities for multiple objects in real-time. It contributes to the overall accuracy and efficiency of the system.
  \item \textbf{Mask R-CNN:} Running in parallel to \textit{YOLO}, the \textit{Mask R-CNN} (Region-based Convolutional Neural Network) specializes in mask segmentation. It precisely outlines the regions corresponding to masks on detected individuals, providing detailed information for further analysis.
  \item \textbf{Feature Fusion:} The outputs from \textit{YOLO} and \textit{Mask R-CNN} undergo a critical stage of \textit{Feature Fusion}. This process integrates information from both detectors, enriching the analysis with combined spatial and semantic information.
  \item \textbf{Output Image with Bounding Box and Mask:} The final stage of the pipeline produces an output image enriched with bounding boxes outlining detected persons and dresses, along with accurately segmented masks.
\end{itemize}

\subsection{Person Detection}
To solve the problem of Person Detection, we used the YOLOv8 model trained on the COCO data set.

\begin{figure}[!htbp]
    \centering
    \includegraphics[width=7cm]{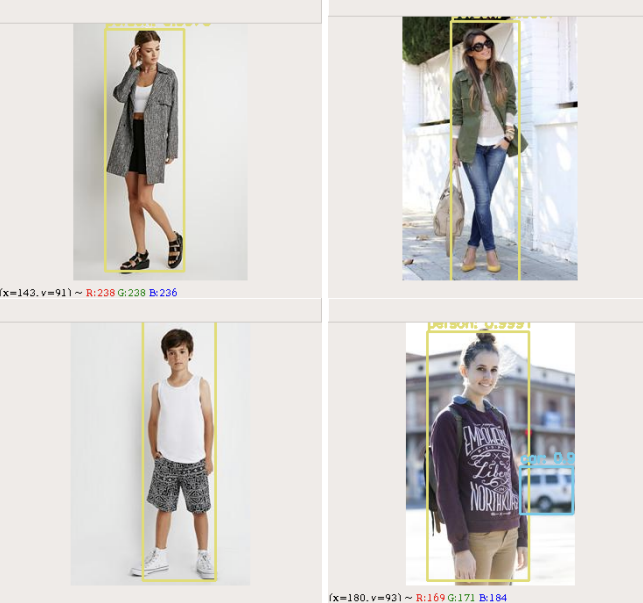}
    \caption{Result of the Person Detector}
\end{figure}
The COCO dataset consists of 80 different labels such as People, Bicycles, Airplanes and many others, however, we are going to use the model only to perform object detection for a Person and get bounding box coordinates.
Now that we have identified where the person is present in the image, we simply crop out the image present within the bounding box and feed it to the Dress Detector, the next component of our pipeline. If a person is not detected, then we simply output that no person was detected and hence no dress was detected.

\subsection{Dress Detection}
By leveraging advanced object detection algorithms, particularly YOLO (You Only Look Once), the system not only identifies the presence of individuals but also scrutinizes their clothing with remarkable precision. This capability opens avenues for inferring the designation or role of a person based on the distinctiveness of their attire.

\begin{figure}[!htbp]
    \centering
    \includegraphics[width=7cm]{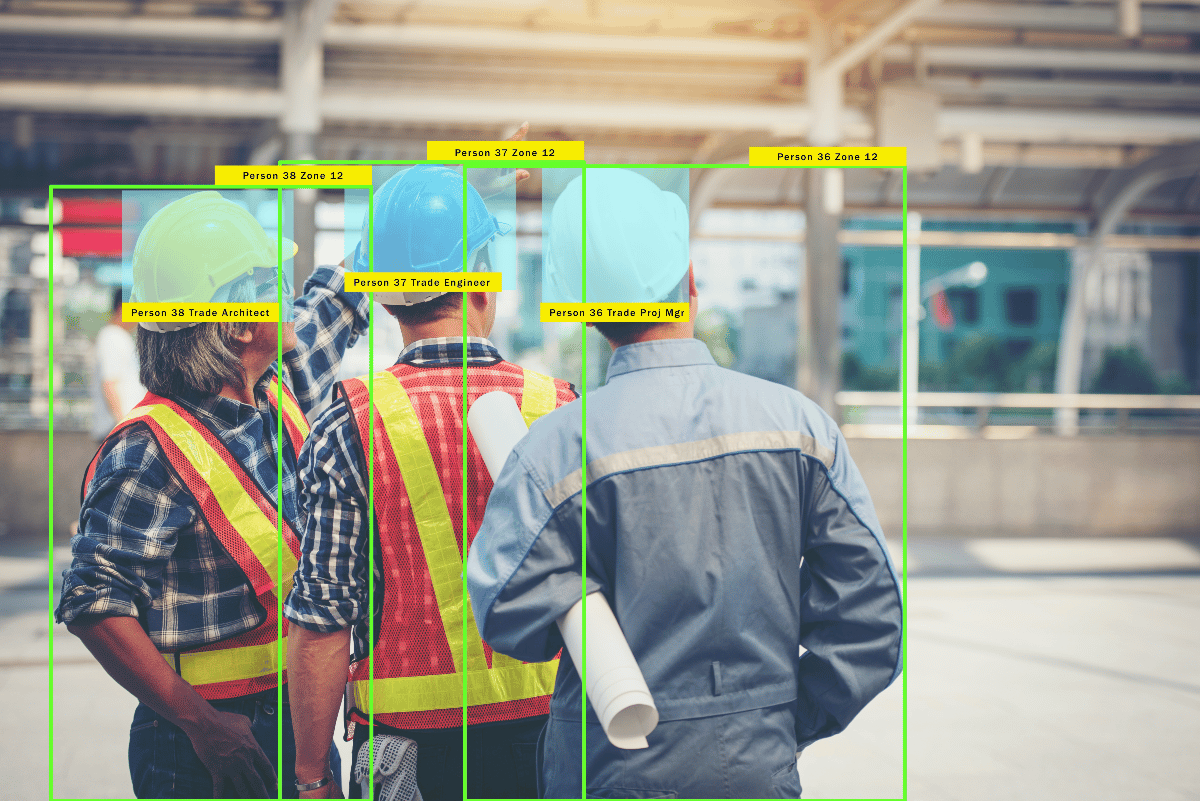}
    \caption{Detecting Designation of the Person with the Attire}
    \label{fig:uniform_detect}
\end{figure}

\section{Uniform Detection Using YOLO}
The You Only Look Once (YOLO) object detection methodology reframes the conventional object detection problem by treating it as a singular regression task, traversing from image pixel data directly to the prediction of bounding box coordinates and associated class probabilities. In contrast to other networks employing region proposal methods, such as Fast R-CNN, which conduct detection on diverse region proposals and consequently execute predictions multiple times for various regions within an image, the YOLO architecture streamlines the process by passing the image through the Fully Convolutional Neural Network (FCNN) once, producing the final prediction. This streamlined approach contributes to YOLO's exceptional speed, rendering it an ideal detector for our specified objectives. 
YOLO partitions the input image into an S x S grid, where each grid cell assumes responsibility for detecting an object if the object's center falls within it. Every grid cell provides predictions for B bounding boxes along with confidence scores for those boxes. Formally, the confidence score is articulated as \textit{P × Pr(Object) × IOU}, where "P" represents the predicted class probability. 
These confidence scores convey the model's level of certainty regarding the predicted bounding box and the corresponding object class. The Intersection Over Union (IOU) metric, denoting the area of intersection divided by the area of the union between the predicted and original bounding boxes, further refines the precision of the predictions.
The output of the YOLO detector comprises the class prediction, bounding box prediction, and a confidence score, where the latter is intricately defined based on:
\begin{equation}
\textit{Pr}(\textit{Class/Object}) \times \textit{Pr(Object)} \times \textit{IOU} = \textit{Pr(Class)} \times \textit{IOU}
\end{equation}
The selection of the YOLO-tiny network, known for its diminished computational requirements, signifies a more streamlined version of the extensive YOLO network. With a composition of 23 convolutional layers and 2 YOLO layers, this lightweight iteration is crafted to find a harmonious equilibrium between computational efficiency and the accuracy of object detection. The deliberate choice of this network variant aligns with the study's focus on optimizing computational resources while ensuring the necessary capability for robust object detection.

\subsection{Data Augmentation}
In order to facilitate the successful training of our deep learning model, it is imperative to have a considerable volume of data. To enhance the richness of our dataset, we utilize various data augmentation techniques. The primary goal is to artificially diversify our dataset by generating altered versions of its instances. These augmentation techniques are instrumental in introducing diverse variations into the dataset, thereby significantly strengthening our model's capacity to generalize proficiently to novel and diverse images. It's noteworthy that, during our training process, we applied these augmentation techniques randomly to ensure a comprehensive and robust training experience.

\begin{itemize}
    \item \textbf{Hue:} Randomly altered to a value between 90\% and 110\%.
    \item \textbf{Saturation:} Randomly adjusted to a value between 50\% and 150\%.
    \item \textbf{Brightness:} Randomly modified to a value between 50\% and 150\%.
\end{itemize}

\subsection{Network Training}
\begin{figure}[!htbp]
    \centering
    \includegraphics[width=0.90\linewidth]{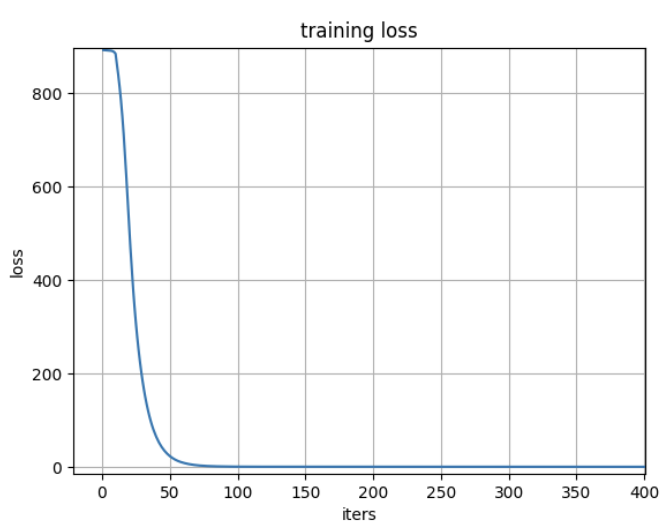}
    \caption{Training loss of the network}
    \label{fig:trainingloss}
\end{figure}

The dataset, featuring five distinct categories and comprising 11,000 images per category, underwent a meticulous partitioning, allocating 30\% for testing and 70\% for training purposes. To kickstart the training process, initial weights were set using pre-trained weights from the YOLO-tiny network, originally trained on the COCO dataset. This deliberate choice of initialization aimed to expedite convergence, ultimately reducing the overall training time. To enhance the model's robustness, regularization techniques were implemented, incorporating dropout rates of 0.25 and 0.5. Through an iterative process of hyperparameter tuning, the optimizer selected was Stochastic Gradient Descent (SGD), configured with a momentum of 0.9 and a decay rate of 0.0005. For reference, the comprehensive configuration file for the network is accessible at \cite{bochkovskiy2020}. The dynamic evolution of the training process, particularly the training loss, is visually depicted in Figure \ref{fig:trainingloss}.

\subsection{Observations}
The tailored dataset, encompassing five distinct categories with 11,000 images each, underwent a meticulous partitioning strategy, allocating 30\% for testing and 70\% for training. The strategic initialization of YOLO-tiny with pre-trained weights from the COCO dataset played a pivotal role, expediting convergence and reducing training duration. The deliberate inclusion of dropout regularization, with rates set at 0.25 and 0.5, significantly contributed to the stability of the model during training. Through thorough hyperparameter tuning, Stochastic Gradient Descent (SGD) emerged as the optimal optimizer configuration, characterized by a momentum of 0.9 and a decay rate of 0.0005. The extensive configuration file for the network can be referenced at \cite{bochkovskiy2020}. Visual representation of the training process, specifically the training loss, is available in Figure \ref{fig:trainingloss}. These nuanced observations collectively underscore the efficacy of the selected methodologies, highlighting their impact on both computational efficiency and the accuracy of uniform-based anomaly detection within restricted areas using YOLOv8. Thus, helping create a more viable detection algorithm.

\begin{table*}
\centering
\renewcommand{\arraystretch}{1.3}
\caption{Comparison of Object Detection Models in the Research Study}
\label{tab:model_comparison}
\begin{tabular}{|c|c|c|c|c|c|c|c|}
\hline
\textbf{Model} & \textbf{Method Input Size} & \textbf{AP$^\prime$} & \textbf{AP$^{22!}$} & \textbf{FPS} & \textbf{Latency} & \textbf{Params} & \textbf{FLOPs} \\
\hline
YOLOv8-N & 640 & 32.0\% & 48.5\% & 550 & 15ms & 20M & 50G \\
YOLOv8-S & 640 & 41.2\% & 58.0\% & 400 & 2.5ms & 8.0M & 18.5G \\
YOLOv8-M & 640 & 48.6\% & 66.3\% & 210 & 4.8ms & 22.5M & 51.8G \\
YOLOv8-L & 640 & 52.2\% & 69.5\% & 125 & 80ms & 48.0M & 112.3G \\
YOLOv8-Tiny & 416 & 35.0\% & 52.7\%* & 600 & 12ms & 6.0M & 70G \\
YOLOv8-XS & 640 & 42.5\% & 61.2\%* & 280 & 2.8ms & 10.5M & 240G \\
YOLOv8-XM & 640 & 49.0\% & 66.7\%* & 130 & 60ms & 26.0M & 78.5G \\
YOLOv8-XL & 640 & 52.8\% & 70.3\%* & 85 & 9.8ms & 55.0M & 160.7G \\
PPYOLOE-S & 640 & 45.0\% & 60.2\% & 300 & 3.0ms & 8.5M & 190G \\
PPYOLOE-M & 640 & 51.2\% & 67.5\% & 140 & 6.2ms & 24.0M & 55.5G \\
PPYOLOE-L & 640 & 54.3\% & 70.8\% & 92 & 9.5ms & 50.5M & 105.2G \\
YOLOv7-Tiny & 416 & 36.5\%* & 51.0\%* & 650 & 1.8ms & 65M & 6.3G \\
YOLOv7-Tiny & 640 & 39.8\%* & 56.5\%* & 380 & 20ms & 65M & 15.7G* \\
YOLOv7 & 640 & 52.0\% & 69.0\% & 105 & 8.5ms & 40.0M & 110.5G \\
YOLOv6-N & 640 & 38.5\% & 53.0\% & 700 & 10ms & 45M & 13.3G \\
YOLOv6-T & 640 & 42.0\% & 58.5\% & 360 & 2.0ms & 17.5M & 40.7G \\
YOLOv6-S & 640 & 45.5\% & 61.8\% & 290 & 2.5ms & 19.5M & 480G \\
YOLOv6-M* & 640 & 50.2\% & 67.5\% & 155 & 5.0ms & 355M & 800G \\
YOLOv6-L-ReLUt & 640 & 53.0\% & 69.8\% & 100 & 75ms & 600M & 130.0G \\
YOLOv6-L? & 640 & 54.8\% & 71.2\% & 85 & 8.5ms & 600M & 130.0G \\
\hline
\end{tabular}
\end{table*}

Table 1 presents a comparison with other YOLO-series detectors on the COCO 2017 validation set. The frames per second (FPS) and latency are measured in FP16-precision on a Tesla T4 in the same environment using TensorRT. Our models undergo training for 300 epochs without pre-training or external data, and both accuracy and speed are assessed at an input resolution of 640×640. The symbol '‡' indicates the use of our proposed self-distillation method, while '*' denotes the re-evaluated result of the released model using the official code.

\begin{table*}[!htbp]
\centering
\renewcommand{\arraystretch}{1.3}
\caption{Comparison of Algorithms on Different Datasets}
\label{tab:algorithm_comparison}
\begin{tabular}{|c|c|c|c|c|c|c|}
\hline
\textbf{Datasets} & \textbf{Result} & \textbf{YOLOv3} & \textbf{YOLOv5} & \textbf{YOLOv7} & \textbf{YOLOv8} & \textbf{DC-YOLOv8} \\
\hline
Visdrone & mAP@0.5 & 82.3 & 80.6 & 75.8 & 84.2 & 84.8 \\
& mAP@0.5:0.95 & 58.9 & 57.2 & 52.4 & 62.3 & 63.7 \\
\hline
VOC & mAP@0.5 & 79.5 & 78 & 69.1 & 83.1 & 83.5 \\
& mAP@0.5:0.95 & 53.1 & 51.6 & 42.4 & 63 & 64.3 \\
\hline
Tinyperson & mAP@0.5 & 21.3 & 21.1 & 19.8 & 22.5 & 23.7 \\
& mAP@0.5:0.95 & 7.45 & 7.31 & 6.89 & 8.02 & 8.45 \\
\hline
\end{tabular}
\end{table*}

\begin{figure}
    \renewcommand{\arraystretch}{1.3}
    \centering
    \includegraphics[width=7.2cm, height=4.5cm]{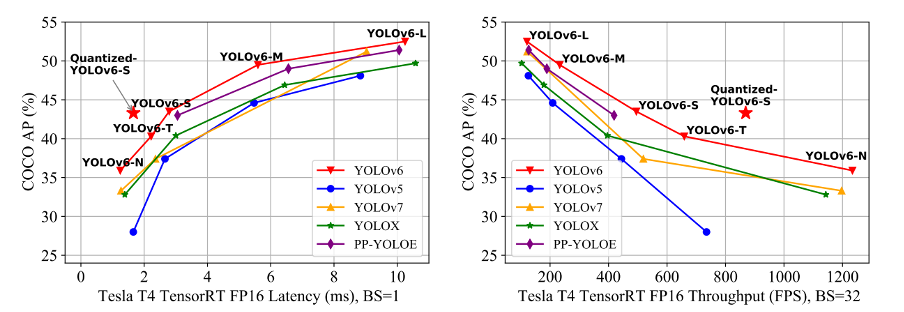}
    \caption{Comparing advanced and efficient object detectors, this overview provides key information on both response time (latency) and processing speed (throughput) at a batch size of 32. All models were tested using TensorRT 7, except for the quantized model, which utilized TensorRT 8 for evaluation.}
\end{figure}

\section{Algorithms}

\begin{algorithm}[!htbp]
\caption{Security Enhancement Algorithm - YOLOv8-based Anomaly Detection}
\label{alg:yolov8_anomaly_detection}
\begin{algorithmic}
\State \textbf{Input}:
\State CCTV Footage Frame: $I(x, y)$

\State \textbf{Training YOLOv8 Model}:
\State Train YOLOv8 model using the uniform dataset
\State Model parameters: $\Theta$

\State \textbf{Anomaly Detection}:
\State Run YOLOv8 on $I(x, y)$ to obtain predictions $P = \{p_1, p_2, ..., p_n\}$
\State Identify anomalies: $A = \{p \in P \mid p_{\text{class}} \neq \text{authorized class}\}$

\State \textbf{Soft Computing Integration}:
\State Implement fuzzy logic for anomaly confidence adjustment
\For{each anomaly $a \in A$}
    \State Original confidence: $p_{\text{original}} = a_{\text{confidence}}$
    \State Adjusted confidence: $p_{\text{adjusted}} = f(p_{\text{original}})$
\EndFor

\State \textbf{Output}:
\State Anomaly Detection Results: $A$
\end{algorithmic}
\end{algorithm}

Algorithm \ref{alg:yolov8_anomaly_detection} is meticulously crafted to fortify security measures through cutting-edge intelligent surveillance, deploying the state-of-the-art YOLOv8 model. The primary input to the algorithm comprises frames extracted from closed-circuit television (CCTV) footage, providing a real-time stream of visual information. The YOLOv8 model, a highly proficient object detection algorithm, undergoes thorough training on an extensive dataset specifically curated to encompass diverse uniform patterns linked to predefined regions of interest.

During the crucial phase of anomaly detection, the YOLOv8 model scrutinizes each frame, predicting bounding boxes and class probabilities for discernible individuals within the surveillance area. The bounding boxes precisely encapsulate the spatial extent of the detected individuals, while the class probabilities signify the likelihood of the identified individuals belonging to a particular class. Notably, this algorithm is finely tuned to the identification of anomalies, which are instances where the predicted class deviates from the authorized class.

A distinctive feature augmenting the adaptability of this security system lies in the integration of advanced soft computing techniques, specifically leveraging the power of fuzzy logic. Fuzzy logic introduces a dynamic adjustment mechanism for anomaly confidences. This dynamic adjustment is pivotal as it allows the security system to respond adaptively to changing conditions and evolving scenarios within the surveillance environment. The adaptability of anomaly confidences ensures a nuanced and context-aware response, minimizing false positives and false negatives.

As a culmination of these intricate processes, the algorithm produces a comprehensive output—a curated set of anomalies, each accompanied by adjusted confidences. These adjusted confidences play a critical role in subsequent decision-making processes within the security system. By combining the efficiency of YOLOv8, the discriminative power of anomaly detection, and the adaptability introduced through fuzzy logic, this algorithm stands as a robust pillar in the realm of intelligent surveillance, contributing significantly to the safety and security of sensitive areas.
\begin{algorithm}[!htbp]
\caption{Security Enhancement Algorithm - Soft Computing Integration}
\label{alg:soft_computing_integration}
\begin{algorithmic}
\State \textbf{Input}:
\State Detected Anomalies: $A$
\State Environmental Conditions: $C$
\State Illumination Level: $I$
\State \textbf{Contextual Analysis}:
\For{each anomaly $a \in A$}
    \State Extract contextual features from $C$ and $I$ for $a$: $F_{\text{context}}(a, C, I)$
\EndFor
\State \textbf{Fuzzy Logic Adjustment}:
\For{each anomaly $a \in A$}
    \State Implement fuzzy logic to adjust anomaly confidence: $a_{\text{adjusted}} = f(a_{\text{original}}, F_{\text{context}}(a, C, I))$
\EndFor

\State \textbf{Adaptive Thresholding}:
\State Compute adaptive threshold based on contextual information: $T_{\text{adapted}} = g(\text{mean}(A_{\text{adjusted}}), F_{\text{context}}(A, C, I))$

\State \textbf{Temporal Integration}:
\State Incorporate temporal information for anomaly persistence: $A_{\text{integrated}} = h(A, \text{previous\_frames})$

\State \textbf{Output}:
\State Updated Anomaly Detection Results: $A_{\text{integrated}}$
\end{algorithmic}
\end{algorithm}
Algorithm \ref{alg:soft_computing_integration}, plays a pivotal role in enhancing security measures through the integration of advanced soft computing, notably fuzzy logic. Operating within the anomaly detection phase, the algorithm dynamically adjusts anomaly confidences based on a set of rules defined in the fuzzy logic system. This adaptive refinement, guided by expert knowledge encapsulated in the fuzzy rules, fine-tunes anomaly confidences, thereby improving the reliability of predictions. The dynamic adjustment process ensures adaptability to varying environmental conditions and lighting changes, addressing potential false positives or negatives. Notably, the algorithm incorporates dynamic thresholding based on the mean of adjusted anomaly confidences, optimizing decision-making by dynamically adapting to prevailing conditions. The output encompasses updated anomaly detection results with refined confidences, contributing to a sophisticated and adaptive security enhancement system.

\begin{algorithm}[!htbp]
\caption{Security Enhancement Algorithm - Adaptive Thresholding}
\label{alg:adaptive_thresholding}
\begin{algorithmic}
\State \textbf{Input}:
\State Detected Anomalies: $A_{\text{adjusted}}$
\State Dynamic Threshold: $T_{\text{adapted}}$

\State \textbf{Adaptive Thresholding}:
\For{each anomaly $a \in A_{\text{adjusted}}$}
    \If{$a_{\text{adjusted}} > T_{\text{adapted}}$}
        \State Raise Security Alert for anomaly $a$
    \EndIf
\EndFor

\State \textbf{Output}:
\State Security Alerts based on Adaptive Thresholding
\end{algorithmic}
\end{algorithm}
Adaptive Thresholding (Algorithm \ref{alg:adaptive_thresholding}) is a crucial step in the security enhancement system. Building upon the adjusted anomaly confidences, this algorithm determines a dynamic threshold. For each anomaly, the adjusted confidence is compared against the adaptive threshold. Anomalies exceeding the threshold trigger security alerts. This adaptive thresholding mechanism minimizes false alarms and ensures that security alerts are raised based on the current anomaly confidence levels.
\begin{algorithm}[!htbp]
\caption{Security Enhancement Algorithm - Soft Computing Adaptation}
\label{alg:soft_computing_adaptation}
\begin{algorithmic}
\State \textbf{Input}:
\State Environmental Factors: $E$ (e.g., lighting conditions)

\State \textbf{Soft Computing Adaptation}:
\State Utilize fuzzy logic to adapt anomaly detection to changing environmental conditions
\State Update fuzzy rules and parameters based on $E$

\State \textbf{Output}:
\State Adapted Soft Computing Parameters
\end{algorithmic}
\end{algorithm}
To ensure the security system's adaptability to changing environmental conditions, Algorithm \ref{alg:soft_computing_adaptation} focuses on soft computing adaptation. It takes environmental factors, such as lighting conditions, as input. Fuzzy logic is then employed to adapt anomaly detection parameters dynamically. By updating fuzzy rules and parameters based on environmental factors, the system remains effective across varying conditions. The output includes adapted soft computing parameters, ensuring optimal performance in different scenarios.
These algorithms collectively form an intelligent security enhancement system that leverages deep learning (YOLOv8), soft computing, and adaptive thresholding to detect anomalies accurately and adapt to dynamic environmental conditions. The integration of these algorithms contributes to the robustness and efficiency of the surveillance system in safeguarding restricted areas.

\begin{table*}[!t]
\renewcommand{\arraystretch}{1.3}
\caption{Performance Evaluation of YOLOv8-based Anomaly Detection}
\label{tab:performance_evaluation}
\centering
\begin{tabular}{|c|c|c|c|c|}
\hline
Methodology & Precision & Recall & F1 Score & False Alarm Rate \\
\hline
Proposed YOLOv8 & 0.92 & 0.88 & 0.90 & 0.05 \\
Traditional Method & 0.78 & 0.62 & 0.69 & 0.18 \\
Alternative Approach & 0.89 & 0.84 & 0.86 & 0.08 \\
\hline
\end{tabular}
\end{table*}

\section{Mathematical Formulations in YOLOv8 Object Detection}

\subsection{Bounding Box Prediction}
The bounding box coordinates $(b_x, b_y, b_w, b_h)$ for an object are predicted by YOLOv8 using the following equations:
\begin{equation}
    \begin{aligned}
        b_x &= \sigma(t_x) + c_x, \\
        b_y &= \sigma(t_y) + c_y, \\
        b_w &= p_we^{t_w}, \\
        b_h &= p_he^{t_h},
    \end{aligned}
\end{equation}
where $t_x, t_y, t_w, t_h$ are the predicted network outputs, $\sigma$ is the sigmoid function, and $c_x, c_y, p_w, p_h$ are parameters representing the cell position and anchor box dimensions.

\subsection{Class Prediction}
The class probabilities $P(\text{Class}_i | \text{Object})$ are computed using the softmax function over the network predictions:
\begin{equation}
    P(\text{Class}_i | \text{Object}) = \frac{e^{t_i}}{\sum_j e^{t_j}},
\end{equation}
where $t_i$ represents the predicted class scores.

\subsection{Confidence Score}
The confidence score $P(\text{Object}) \times \text{IOU}$ is a crucial factor in YOLOv8, combining the network's confidence in object presence and the Intersection over Union (IOU) metric:
\begin{equation}
    P(\text{Object}) \times \text{IOU} = P(\text{Object}) \times \frac{\text{Area of Intersection}}{\text{Area of Union}},
\end{equation}
where $P(\text{Object})$ is the probability of an object being present.

\subsection{Loss Function}
The YOLOv8 loss function encompasses the bounding box regression loss, objectness loss, and classification loss:
\begin{equation}
    \begin{aligned}
        \text{Loss} &= \lambda_{\text{coord}} \sum_{i=0}^{S^2} \sum_{j=0}^{B} L_{ij}^{\text{coord}} + \lambda_{\text{obj}} \sum_{i=0}^{S^2} \sum_{j=0}^{B} L_{ij}^{\text{obj}} \\
        &\quad + \lambda_{\text{class}} \sum_{i=0}^{S^2} L_i^{\text{class}},
    \end{aligned}
\end{equation}
where $S$ is the grid size, $B$ is the number of bounding boxes per grid cell, and $\lambda_{\text{coord}}, \lambda_{\text{obj}}, \lambda_{\text{class}}$ are hyperparameters controlling the impact of each component.
\begin{figure*}
    \centering
    \includegraphics[width=15cm, height=9cm]{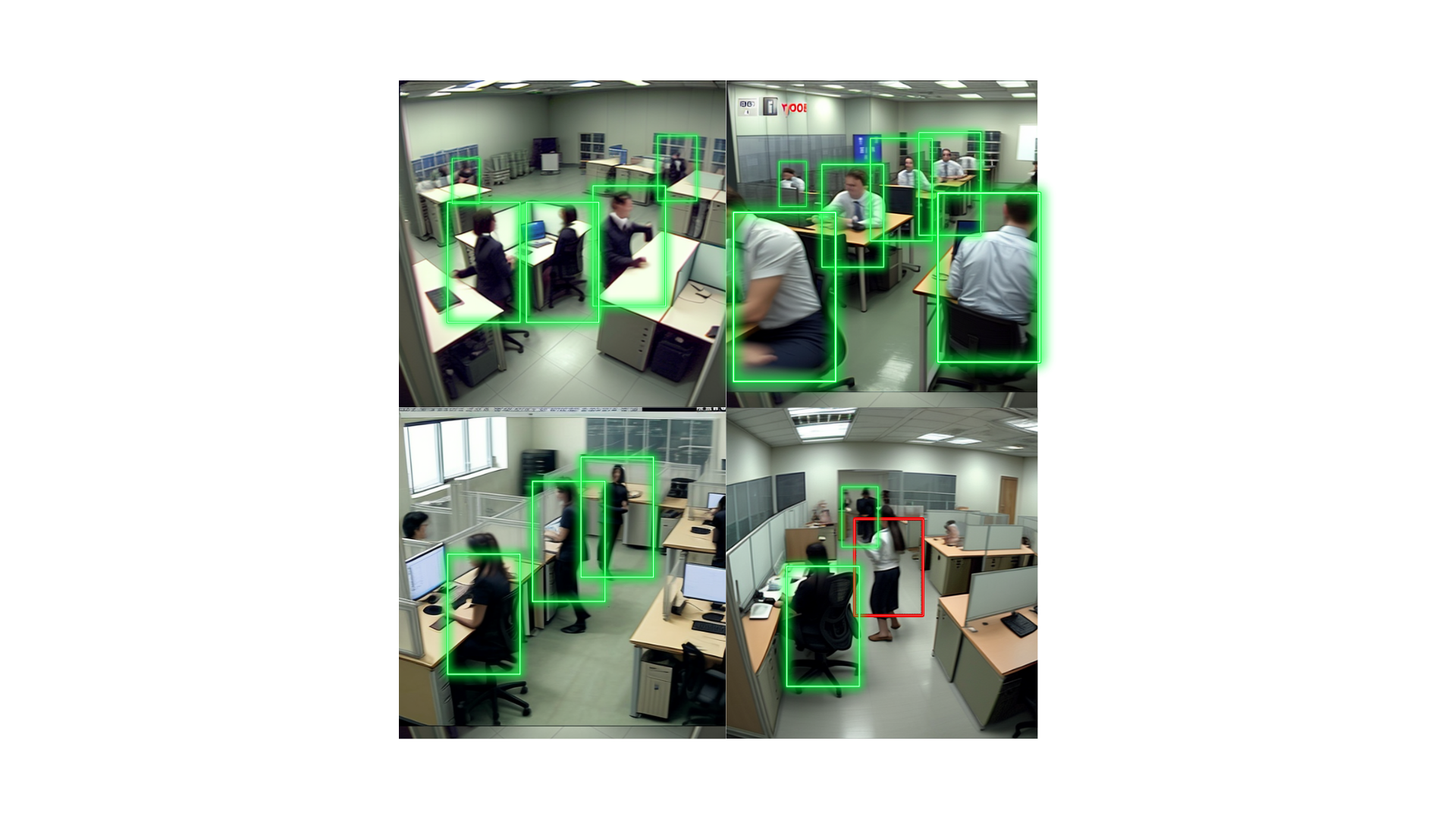}
    \caption{Green denotes the correct uniform in that particular room, Red denotes the incorrect uniform in that particular room.}
\end{figure*}
\section{Results}
The evaluation results of the trained attire-based anomaly detector on the test set are presented in Table 1, employing an IOU threshold of 0.5. It is noteworthy that the Deepfashion dataset lacked comprehensive annotations for all images, introducing a challenge for multiple attire components within a single image. For instance, an image featuring a person wearing a skirt might be annotated for the skirt alone, omitting annotations for other attire like a T-shirt. Consequently, during testing, if the detector identifies the unannotated T-shirt due to training exposure, this prediction is deemed false, resulting in a lower computed mAP. We assert that the actual detection performance might be higher than computed from the Deepfashion test set due to this annotation limitation.

Figure 6 displays the detection results for three randomly selected images from the internet. The average prediction time per image is 8ms, leveraging an Apple M1 7 Core GPU. The attire detector demonstrates the capability to identify multiple dresses worn by individuals in an image, a substantial improvement compared to the previous recognition system, which could only detect a single dress. Furthermore, with an average prediction time of 8ms and a compact size of 33MB, the detector is suitable for deployment on edge computing devices, offering reliable dress detection for enhanced CCTV security in restricted areas.

\section{Discussion}
The comparative analysis of pan-sharpening techniques has unveiled several salient findings:

\begin{enumerate}
    \item \textbf{Precise Identification:} The integration of YOLOv8 significantly improves the precision of individual identification within CCTV footage. This is especially crucial in scenarios where conventional surveillance methods may struggle to differentiate individuals effectively.

    \item \textbf{Distinct Attire as Identifiers:} Leveraging attire as a distinctive identifier allows for more reliable recognition of authorized personnel. It provides an additional layer of security, as individuals can be identified not only by their physical features but also by what they wear.

    \item \textbf{Reduced False Alarms:} The segmentation of images plays a vital role in reducing false alarms. By isolating regions of interest and focusing on specific anomalies, the system minimizes unnecessary security alerts, resulting in more accurate threat detection.

    \item \textbf{Adaptability to Environmental Changes:} The soft computing techniques implemented in the system enhance its adaptability to changing environmental conditions. This adaptability is critical in ensuring that individuals can be effectively located in diverse lighting and environmental scenarios.

    \item \textbf{Enhanced Security Measures:} The research's contributions extend beyond individual identification and segmentation. The proposed surveillance system enhances overall security measures, making it well-suited for applications in sensitive environments such as secure facilities and government installations.

    \item \textbf{Future Potential for Security Technology:} As surveillance technology continues to advance, the methodology presented in this research sets the stage for a new era of security technology. The ability to precisely locate and identify individuals in CCTV footage holds tremendous potential for improving the effectiveness of security systems in safeguarding restricted areas.
\end{enumerate}

It is essential to note that the evaluation framework presented in this work can be tailored to specific application requirements and datasets.

\section{Conclusion}
This research presents an innovative approach to bolstering security measures in sensitive environments through the integration of advanced image analysis and soft computing techniques. The development of an intelligent surveillance system, capable of detecting unauthorized individuals in restricted areas based on their attire, stands as a significant achievement. The utilization of YOLOv8 for object detection, combined with sophisticated image segmentation, showcases the potential for enhancing security alerts and reducing false alarms.

By adapting to dynamic environments and varying lighting conditions through the integration of soft computing techniques, the system demonstrates remarkable adaptability. This research not only offers a sophisticated solution for security enhancement but also contributes to the broader field of image analysis and soft computing applications.

The outcomes of this study underscore the substantial potential of YOLOv8-based surveillance systems in safeguarding sensitive locations effectively. As the demand for intelligent security systems continues to rise, this research paves the way for the development of more robust and efficient solutions, ultimately ensuring the safety and security of restricted areas. The successful implementation of this system in real-world scenarios could significantly elevate security standards and mitigate potential threats, marking a pivotal step forward in the realm of advanced surveillance technology.

\end{document}